%% file: main.tex
\documentclass[runningheads]{llncs}

\usepackage{eccv}

\usepackage{eccvabbrv}

\usepackage{graphicx}
\usepackage{booktabs}
\usepackage{multirow}
\usepackage{algorithm}
\usepackage{algpseudocode}
\usepackage{tikz}
\usepackage{placeins}
\usetikzlibrary{positioning,arrows.meta,calc,patterns,decorations.pathreplacing,fit}
\usepackage[accsupp]{axessibility}  

\usepackage{hyperref}

\usepackage{orcidlink}

\graphicspath{{figures/}}

\begin{document}

\title{AVQ-Attention: Adaptive Vector-Quantized Attention}


\author{Winfried van den Dool\inst{1,2} \and
Patrick Forr\'{e}\inst{3,4} \and
Amir Habibian\inst{5} \and
Yuki M.~Asano\inst{6} \and
Max Welling\inst{2}}

\authorrunning{W.~van den Dool et al.}

\institute{QUVA Lab, University of Amsterdam, The Netherlands \\
\email{w.v.s.o.vandendool@uva.nl} \and
AMLab, Informatics Institute, University of Amsterdam, The Netherlands \and
AI4Science Lab, University of Amsterdam, The Netherlands \and
Korteweg-de Vries Institute for Mathematics, University of Amsterdam, The Netherlands \and
Qualcomm AI Research, Amsterdam, The Netherlands \and
FunAI Lab, University of Technology Nuremberg, Germany}

\maketitle

\begin{abstract}
The $\mathcal{O}(N^2)$ complexity of attention over $N$ tokens remains a computational bottleneck in transformer models. Vector-Quantized (VQ) attention reduces this to $\mathcal{O}(MN)$ by representing keys with $M$ codewords, but applies uniform codebook capacity regardless of where attention mass concentrates: high-attention regions of key space may be coarsely approximated while low-attention regions waste representational capacity. We propose Adaptive Vector-Quantized (AVQ) Attention, which adaptively allocates codebook capacity based on attention importance. Starting from a small set of codewords, our method identifies the most important codes during the forward pass and refines them with pre-learned child codewords, achieving fine-grained quantization where it matters most while maintaining coarse quantization elsewhere.
We develop an implementation using custom Triton kernels that enables the full adaptive refinement process, including importance scoring, child codeword insertion, and parent contribution replacement, to be carried out within the tiled computation paradigm of Flash Attention with minimal overhead. Our approach maintains $\mathcal{O}(MN)$ complexity while achieving improved accuracy-efficiency trade-offs compared to fixed-codebook VQ-attention.
\end{abstract}

\input{sections/introduction}
\input{sections/related_work}
\input{sections/preliminaries}
\input{sections/method}
\input{sections/experiments}
\input{sections/conclusion}

\section*{Acknowledgements}
This work is financially supported by Qualcomm Technologies Inc., the University of Amsterdam, and the allowance Top consortia for Knowledge and Innovation from the Netherlands Ministry of Economic Affairs and Climate Policy.

\bibliographystyle{splncs04}
\bibliography{main}

\newpage
\appendix
\input{sections/appendix}

\end{document}

%% file: sections/introduction.tex
\section{Introduction}
\label{sec:introduction}

\begin{figure}[t]
\centering
\includegraphics[width=1.01\linewidth]{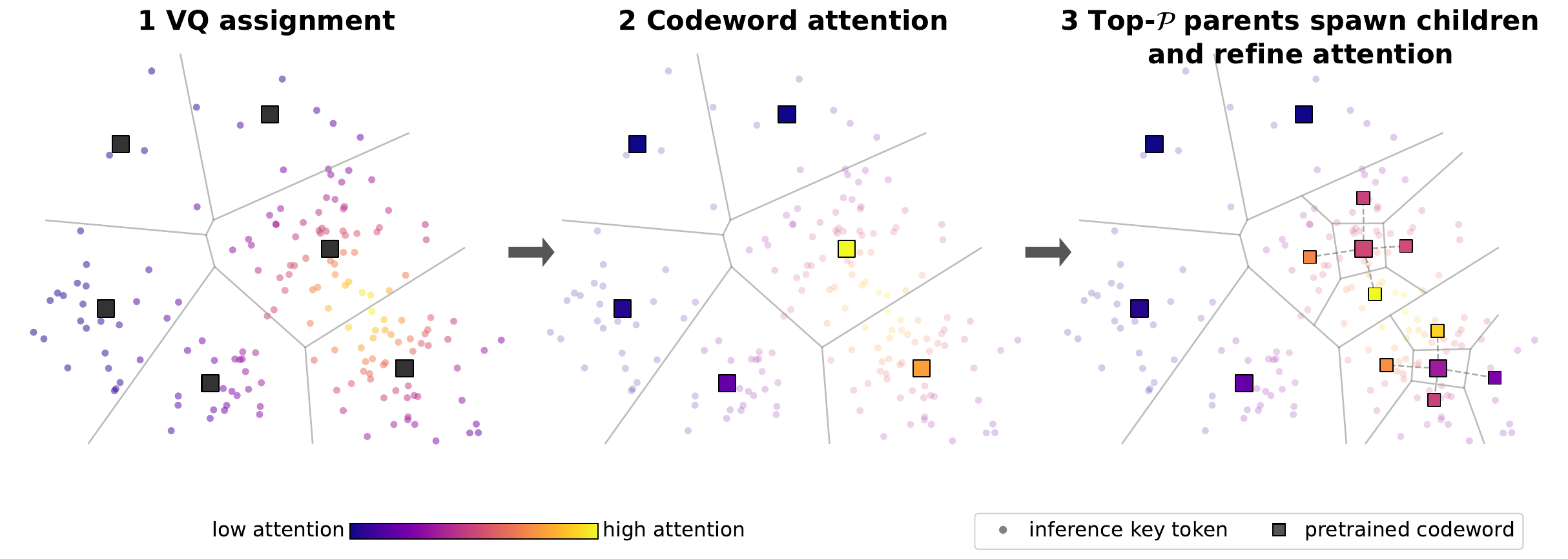}
\caption{key-space visualization of Adaptive VQ-Attention. Small dots represent inference key tokens, colored by how much attention they receive from a given query block; large squares represent pre-trained codewords. \textbf{(1)}~Keys are assigned to their nearest codeword via vector quantization. \textbf{(2)}~Per-codeword importance is computed from the attention mass each codeword receives. \textbf{(3)}~The top-$\mathcal{P}$ most important parents are selected and their pre-learned children are spawned, refining codebook resolution in high-attention regions while leaving low-attention regions at the coarser parent level.}
\label{fig:method_overview}
\end{figure}

Attention mechanisms have become the cornerstone of modern deep learning, enabling transformers \cite{vaswani2017attention} to capture rich interactions between input tokens. However, this expressiveness comes at a computational cost: computing attention between all token pairs requires $\mathcal{O}(N^2)$ operations for sequences of length $N$, making long-sequence processing increasingly prohibitive as models and datasets scale.
Flash Attention \cite{dao2022flashattention} addressed the $\mathcal{O}(N^2)$ memory bandwidth bottleneck through tiling strategies that keep intermediate results in fast SRAM. While this speeds up training and inference on longer sequences, the fundamental computational complexity remains quadratic. Unlike memory traffic, reducing computational cost without approximation is fundamentally impossible: $N$ unique tokens communicating pairwise inherently generate $N^2$ interactions. The challenge therefore becomes how to best trade approximation error for computational efficiency. 

Various approaches have explored this trade-off, from sparse attention patterns that limit which tokens interact \cite{beltagy2020longformerlongdocumenttransformer,child2019generatinglongsequencessparse,roy2021efficient} to token merging methods that reduce the number of tokens processed \cite{bolya2022tokenmerging}. These methods generally modify the transformer's structure by dropping or restricting token interactions.
Vector-quantized attention offers an alternative by clustering keys into $M$ representative codewords where $M<N$ implies reducing complexity to $\mathcal{O}(MN)$. This approach has shown promise \cite{lingletransformer}, but introduces a new challenge: the codebook must be chosen in advance, applying uniform quantization quality across all regions of the key space. During inference, regions receiving high attention mass may suffer from coarse approximation, while low-attention regions waste representational capacity on unnecessary precision. This challenge parallels a well-studied problem in quantization research. Traditional adaptive quantization methods allocate more representational capacity, i.e. bitwidth, to important features while applying coarser quantization elsewhere. These techniques have proven effective across various domains, from image compression to neural network quantization, by concentrating limited resources where they matter most \cite{wallace1991jpeg,wang2019haq,vandendool2025adaptivemesh}.

Inspired by adaptive quantization techniques, we propose Adaptive Vector Quantized (AVQ) Attention. Attention naturally provides an importance signal: by measuring how much attention mass each cluster in key space receives during the forward pass, we obtain a direct importance measure—without the need to design one separately, as in standard adaptive quantization. We use this signal to dynamically allocate codebook capacity where it matters most. Concretely, we start from a set of parent codewords and compute VQ-Attention while extracting per-codeword importance scores. The top-$\mathcal{P}$ most important parents are then refined with their pre-learned child codewords, creating finer quantization in high-attention regions of key space while leaving low-attention regions at the coarser parent level. This refinement adapts dynamically to each input, concentrating codebook capacity based on where that specific input's attention mass falls (see \cref{fig:method_overview}).

Implementing this adaptive refinement efficiently within Flash Attention's tiled framework is non-trivial: the attention weights needed to determine per-codeword importance scores are never materialized---they only exist momentarily at the tile level and are immediately consumed. However, we show that Flash Attention's incremental computation can in fact synergize with our approach: just as Flash Attention builds up attention over blocks of keys via online softmax, AVQ-Attention first computes attention over parent codewords, then incrementally refines it with blocks of children. A geometric constraint on the codebook---each parent equals the mean of its children---allows parent contributions to be recovered directly from the child logits already being computed, enabling efficient in-register correction without revisiting parent codewords. We show that this maintains $\mathcal{O}(MN)$ complexity matching standard VQ-attention while enabling adaptive allocation.

We experimentally validate AVQ-Attention on image classification (ImageNet-1k), semantic segmentation (ADE20K), and high-resolution image generation (Stable Diffusion). We demonstrate improved accuracy-efficiency trade-offs over fixed-codebook VQ-attention and competitiveness with a range of existing efficient-attention methods. Moreover, we show that (A)VQ-attention can be applied post-hoc to pretrained transformers and fine-tuned in a small number of epochs, analogous to quantization-aware training for model compression.
In summary, our contributions include the following.
\begin{itemize}
\item A hierarchical VQ-codebook with a constrained parent-child structure, and a training procedure that learns the codebook end-to-end.
\item A Flash Attention-compatible mechanism that efficiently increments refinements of the attention output.
\item Custom Triton \cite{tillet2019triton} kernels that improve VQ-attention wall-clock performance by fusing operations and minimizing memory traffic, applicable to both flat and adaptive codebooks.
\item A linear complexity attention variant combining these techniques to dynamically allocate compute capacity based on attention importance, adapting to each input at inference time. We validate this experimentally, demonstrating improved accuracy-efficiency trade-offs compared to fixed-codebook VQ-attention.
\end{itemize}

%% file: sections/related_work.tex
\section{Related Work}
\label{sec:related_work}

\paragraph{Adaptive quantization}
allocates representational capacity unevenly based on importance, applying higher precision where it matters most \cite{wallace1991jpeg,wang2019haq}. While these methods typically require a separate mechanism to estimate importance, in the attention setting we can use attention weights directly, alleviating the need for sensitivity analysis \cite{wang2019haq}, learned metrics \cite{tang2022mixedprecision}, or auxiliary models \cite{vandendool2025adaptivemesh}. This insight has also been used to guide mixed-precision KV-cache quantization, allocating higher bitwidth to tokens receiving more attention \cite{zhang2023h2o}. However, adaptivity in the context of VQ-attention can be significantly more effective: rather than varying the precision of $N$ token interactions, VQ-attention reduces the number of interactions itself from $N$ to $M$ codewords, directly addressing the complexity bottleneck.
\paragraph{Vector-Quantization}
has been explored as a means to reduce the quadratic complexity of attention by clustering keys into a smaller set of representative codewords. By attending over codewords rather than individual keys, vector-quantized (VQ) attention reduces computational complexity from $\mathcal{O}(N^2)$ to $\mathcal{O}(MN)$, where $M \ll N$ is the codebook size. Prior work on VQ-attention \cite{lingletransformer} demonstrates that this approach can achieve favorable efficiency–accuracy trade-offs, but relies on a fixed codebook that applies uniform quantization quality across the key space. We build most directly on this line of work, extending VQ-attention in two directions. First, we implement VQ-attention using custom Triton kernels compatible with Flash Attention's tiled computation, minimizing memory traffic. 
Prior VQ-attention work targets the long-context regime where $N \gg M$ and the $\mathcal{O}(MN) \ll \mathcal{O}(N^2)$ complexity gap alone provides large speedups. We first identify opportunities to fuse the sequential steps of VQ-attention into kernels that keep memory bandwidth low, widening the gap with standard attention at large $N$ as the codebook grows. Additionally, it makes VQ-attention competitive already at moderate sequence lengths where the complexity gap is less pronounced and memory bandwidth plays a more significant role. Second, and more centrally, we replace the fixed codebook with an adaptive one that dynamically allocates additional codewords to regions of the key space receiving high attention mass, improving approximation quality where it matters most.
\paragraph{Clustered attention}
approximates full attention by grouping queries into clusters and computing attention once per cluster centroid, achieving linear complexity \cite{vyas2020fast}. This assumes queries within a cluster produce similar attention patterns. Our approach clusters keys rather than queries, so every query retains its own attention pattern over the compressed codeword set. Token merging (ToMe) merges similar tokens into aggregated representations, either permanently reducing the token count and compounding the approximation across layers \cite{bolya2022tokenmerging}, or requiring per-block unmerging to restore the full token set \cite{bolya2023tomesd}. Moreover, finding merge candidates requires computing pairwise token similarities, which is itself quadratic.
\paragraph{Sparse attention}
methods reduce the quadratic cost of attention by restricting computation to a subset of token--token interactions, whether through fixed patterns \cite{beltagy2020longformerlongdocumenttransformer,child2019generatinglongsequencessparse}, content-based routing \cite{roy2021efficient}, locality-sensitive hashing \cite{kitaev2020reformer}, attention-derived selection \cite{yuan2025native}, KV-cache eviction \cite{li2024snapkv}, or hierarchical key selection for long-context inference \cite{hooper2025squeezed,mao2026icecache}.
In contrast, our method preserves dense attention semantics: all keys contribute through their codeword, effectively replacing a binary keep-or-discard decision with a graduated one where less important regions receive coarser approximation rather than being removed entirely.
\paragraph{Linear and low-rank attention}
methods replace the softmax kernel with a decomposable feature map, enabling $\mathcal{O}(N)$ complexity via associativity of matrix multiplication \cite{katharopoulos2020linear}. Low-rank methods such as Linformer \cite{wang2020linformer} project keys and values to a lower-dimensional space. Both families modify the attention mechanism itself, whereas our approach preserves standard softmax attention.

%% file: sections/preliminaries.tex
\section{Preliminaries}
\label{sec:preliminaries}

\subsection{Self-Attention}

Standard scaled dot-product attention \cite{vaswani2017attention} computes outputs as weighted averages of value vectors, where weights are determined by query-key similarities:
\begin{equation}
\label{eq:standardattention}
Y_i = \frac{\sum_{j=1}^{N} \exp(Q_i K_j^{\top}) V_j}{\sum_{l=1}^{N} \exp(Q_i K_l^{\top})}
\end{equation}
where $Q_i, K_j, V_j \in \mathbb{R}^d$ are query, key, and value vectors for tokens $i$ and $j$. This formulation requires $\mathcal{O}(N^2d)$ operations to compute all query-key dot products. It is common to scale key-query dot products by $1/\sqrt{d}$; we omit this for clarity.

\subsection{Vector-Quantized Attention}
VQ-attention reduces complexity by replacing keys with quantized representations from a codebook $\{C_a\}_{a=1}^{M}$ where $M \ll N$. Each key is assigned to its nearest codeword (with slight abuse of notation, we use $a$ both as the assignment function and as a codeword index):
\begin{equation}
\hat{K}_j = C_{a(K_j)}, \quad \text{where} \quad a(K_j) = \arg\min_{a} \|C_a - K_j\|^2
\end{equation}
We can rewrite attention by grouping keys that map to the same codeword. Define:
\begin{equation}
\label{eq:aggregates}
n_a = |\{j : a(K_j) = a\}|, \quad \bar{V}_a = \sum_{j : a(K_j)=a} V_j
\end{equation}
as the count and total aggregated value for codeword $a$. Then
\begin{align}
Y_i &\approx \frac{\sum_{a=1}^{M} \exp(Q_i C_a^{\top}) \bar{V}_a}{\sum_{a=1}^{M} \exp(Q_i C_a^{\top}) n_a},
\end{align}
where the approximation error comes solely from key quantization. Indeed, the equivalence to quantized-key attention follows from regrouping the sums:
\begin{align*}
\sum_{j} \exp(Q_i K_j^\top)V_j &\approx \sum_{j} \exp(Q_i \hat{K}_j^\top)V_j \\
&= \sum_a \sum_{j : a(K_j)=a} \exp(Q_i C_a^\top)V_j \\
&= \sum_a \exp(Q_i C_a^\top) \bar{V}_a
\end{align*}
and similarly for the denominator.

\subsection{Flash Attention}
\label{sec:flashattention}
Flash Attention \cite{dao2022flashattention,dao2024flashattention2} computes attention without materializing the full $N \times N$ matrix $\exp(QK^{\top})$, instead processing tiles that fit in on-chip SRAM. The algorithm maintains running numerators and denominators, scaled by a running maximum for numerical stability \cite{milakov2018online}. As we will see, this incremental structure synergizes naturally with our adaptive refinement. We adopt tile-level notation for the following sections: $I$ and $J$ denote contiguous index sets for query and key (or codeword) tiles respectively, with each tile assumed to fit in SRAM. Rewriting \cref{eq:standardattention}, the actual computation on hardware more closely resembles:
\begin{eqnarray*}
A_{IJ} &=& \exp(Q_I^{} K_J^{\top}) \\
X_I(J) &=& A_{IJ} V_J \\
Z_I(J) &=& \sum_{j \in J} A_{Ij} \\
Y_I &=& \frac{\sum_J X_I(J)}{\sum_J Z_I(J)}.
\end{eqnarray*}

%% file: sections/method.tex
\section{Method}
\label{sec:method}

\subsection{Hierarchical Codebook}

We describe the single-head, single-layer case; multi-head attention maintains one codebook per head. The codebook consists of $M$ parent codewords $\{C_j\}_{j=1}^{M}$, each with $\mathcal{C}$ children $\{C_{j,c}\}_{c=1}^{\mathcal{C}}$, giving $M(1 + \mathcal{C})$ codewords in total, as children supplement rather than replace their parents.

\begin{figure}[t!]
\centering
\includegraphics[width=\linewidth]{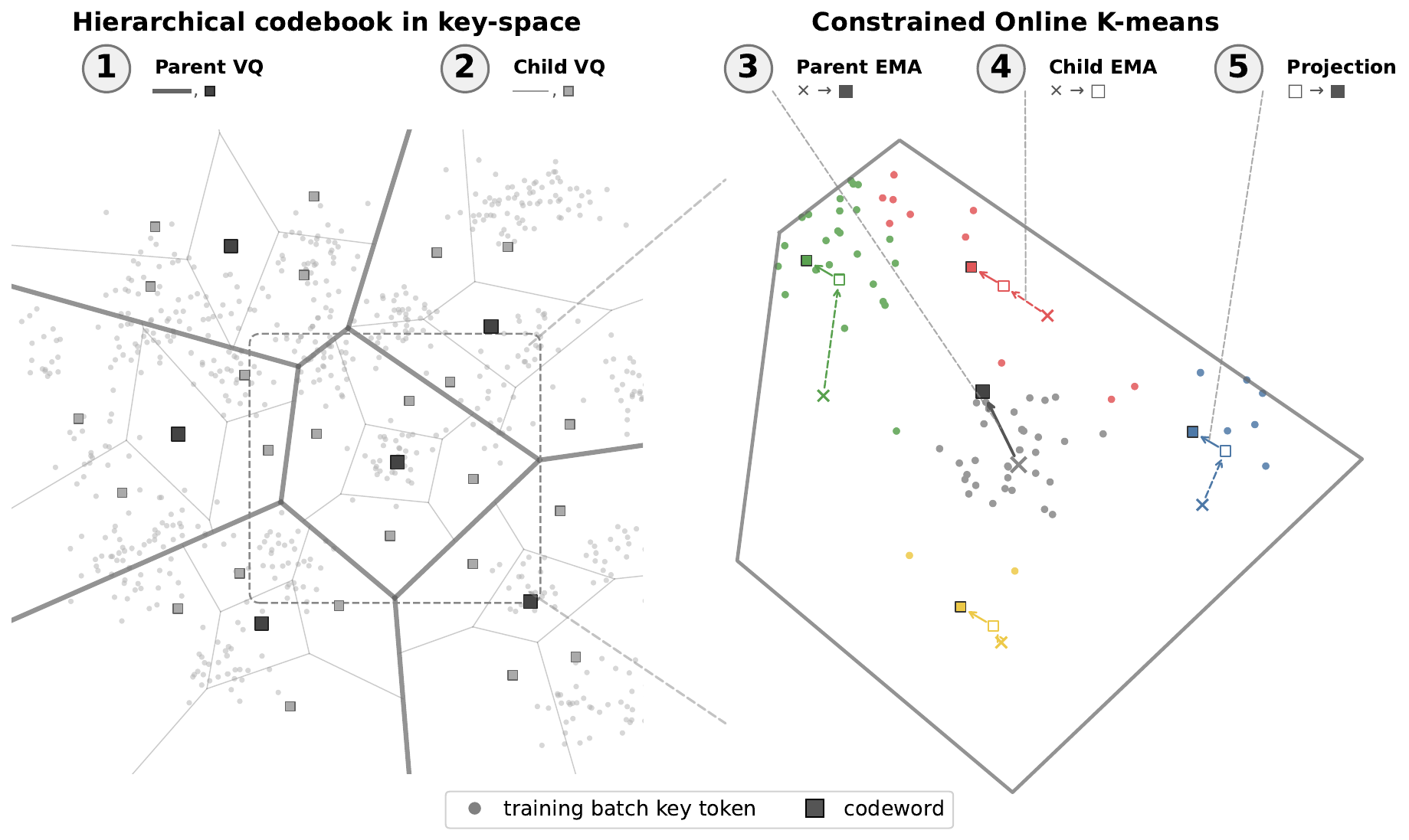}
\caption{Hierarchical codebook in key space, illustrating the five online learning steps.
\textbf{Left:} An incoming training batch (gray dots) arrives at the existing codebook.
\textcircled{\small 1}~Keys are assigned to the nearest parent codeword (thick Voronoi boundaries), then
\textcircled{\small 2}~further quantized to child codewords within each parent's cell (thin boundaries).
\textbf{Right:} Zoom into one parent's cell showing the codebook update.
\textcircled{\small 3}~The parent updates its position via EMA over all keys in its cell.
\textcircled{\small 4}~Each child updates independently via count-weighted EMA.
\textcircled{\small 5}~Children are projected to restore the constraint $C_p = \frac{1}{\mathcal{C}}\sum_c C_{p,c}$; heavier children (larger~$n_c$) resist displacement more (\cref{app:codewordtraining}).}
\label{fig:codebook_keyspace}
\end{figure}

\paragraph{Training.} All codewords are learned via online k-means \cite{lloyd1982least} with exponential moving averages (EMA) \cite{vanoord2017vqvae}. At each training step, keys are first quantized to their nearest parent, and parent centroids are updated via EMA. Each parent's assigned keys are then further quantized among its $\mathcal{C}$ children, with the parent itself remaining as an option---keys that are already well-represented by the parent need not move to a child. Child centroids are updated via EMA on their respectively assigned keys (see \cref{fig:codebook_keyspace}).

\paragraph{Parent-child constraint.} We impose the constraint that each parent equals the mean of its children:
\begin{equation}
\label{eq:parent_constraint}
C_p = \frac{1}{\mathcal{C}}\sum_{c=1}^{\mathcal{C}} C_{p,c}
\end{equation}
Since the unconstrained EMA updates on children will generally violate this, we project child positions back onto the constraint surface after each update via a closed-form mass-weighted projection (see \cref{app:codewordtraining}). This constraint is central to our method: it enables efficient removal of parent attention contributions at inference time, as we show in \cref{sec:child_attention}.

At inference, all codeword positions are fixed.

\subsection{Vector Quantization for (A)VQ-Attention}

Before computing attention, each key must be assigned to a codeword. Unlike standard vector quantization, which only requires assignments, the vector quantization step in VQ-attention must additionally aggregate the values and counts per codeword ($\bar{V}_a$ and $n_a$ in \cref{eq:aggregates}), since these serve as the inputs to the subsequent attention computation. We compute both assignments and aggregates for the full codebook tree in a single fused kernel pass over the keys. Each key is first assigned to its nearest parent among $M_0$ root codewords. Then, it is compared against only the $\mathcal{C}$ children of its assigned parent, and reassigned if a child is closer than the (cached) parent distance. At each level, the key's value is scattered to the assigned codeword, accumulating $\bar{V}_a$ and $n_a$.

The tree structure makes this efficient: the cost per key is $\mathcal{O}(M_0 + \mathcal{C})$ distance computations, yielding a codebook with $M_{\text{total}} = M_0(1 + \mathcal{C})$ codewords.
One may be tempted to defer the computation of child codeword aggregates until after the most important parents have been identified, computing only for children that will actually be used. However, precomputing the full tree brings two key advantages:
\begin{figure}[t!]
\centering
\includegraphics[width=\linewidth]{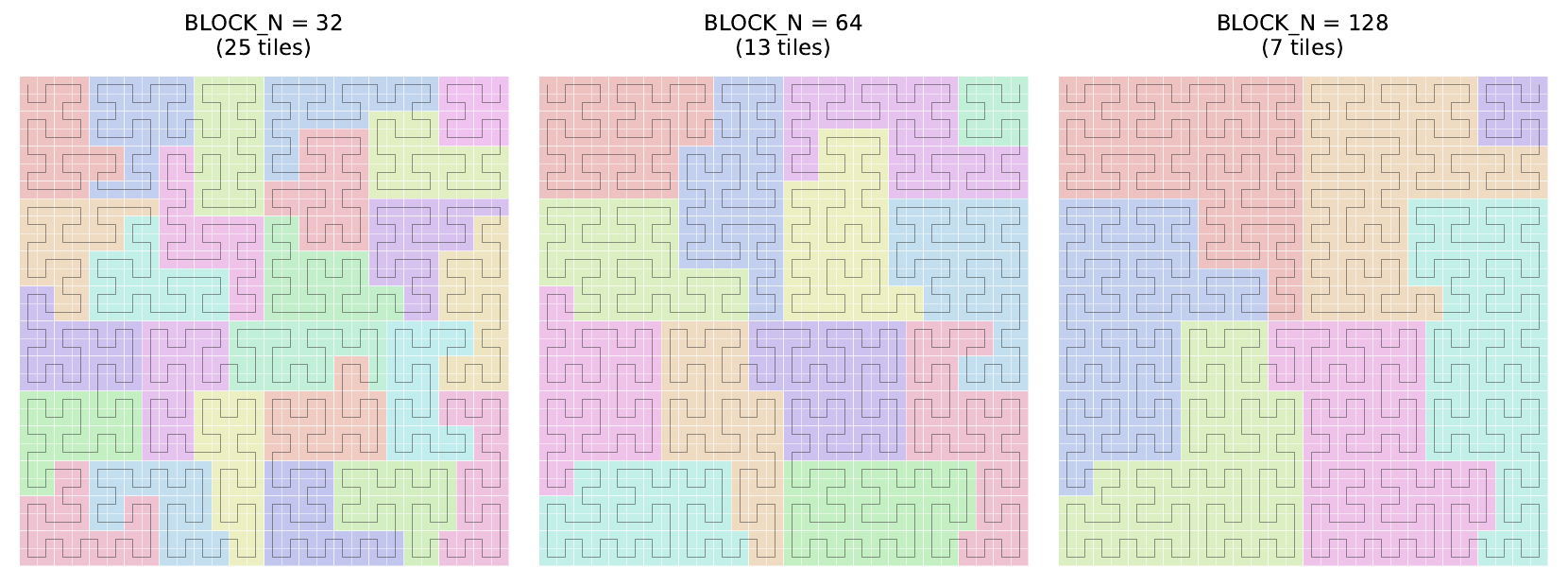}
\caption{Spatial locality of query tiles under Gilbert curve reordering \cite{cerveny2018gilbert} on a $28{\times}28$ patch grid. Tokens are reordered along a Gilbert space-filling curve (gray line) so that contiguous tiles (colored regions) form spatially compact 2D regions, regardless of tile size. Since each tile independently selects which parents to refine, spatial compactness ensures that queries sharing a refinement decision attend to similar regions.}
\label{fig:pertile_adaptivity}
\end{figure}

\begin{itemize}
\item \textbf{Reduced HBM traffic.} The subsequent attention kernel can remain fully fused: after computing attention over parent codewords and determining importance, child aggregates are immediately available, and the kernel can proceed to refine without interrupting to perform additional vector quantization. This avoids writing and re-reading intermediate accumulators to global memory.
\item \textbf{Per-query-tile amortization and adaptivity}. The precomputed aggregates are shared across all query tiles, amortizing the VQ cost. Each tile may then independently select different parents for refinement, making the method adaptive not only per input, but per query tile. To ensure that queries within a tile attend to similar spatial regions, we reorder tokens along a Gilbert space-filling curve \cite{cerveny2018gilbert}, a generalization of the Hilbert curve to non-power-of-two grids. This produces spatially compact tiles regardless of tile size, allowing different regions of the input to refine different parts of the codebook (see \cref{fig:pertile_adaptivity}, \cref{app:gilbert_ablation}).
\end{itemize}
\newpage
\renewcommand{\algorithmicrequire}{\textbf{Input:}}
\renewcommand{\algorithmicensure}{\textbf{Output:}}
\begin{algorithm}[ht!]
\caption{VQ Precompute kernel.}
\label{alg:vqprecompute}
\begin{algorithmic}[1]
\Require $K, V \in \mathbb{R}^{BH \times N \times D}$, \; $C \in \mathbb{R}^{H \times M_{\text{total}} \times D}$ with $M_{\text{total}} = M_0(1+\mathcal{C})$
\Ensure $\bar{V} \in \mathbb{R}^{BH \times M_{\text{total}} \times D}$, \; $n \in \mathbb{R}^{BH \times M_{\text{total}}}$
\State Initialize $\bar{V} = 0$, $n = 0$
\For{each $(bh, \text{Block}_N)$ \textbf{in parallel}} \Comment{Separate GPU programs}
    \State $\mathbf{k} = K[bh, \text{Block}_N]$ \Comment{$[\text{Block}_N, D]$, stays in registers}
    \State $\mathbf{v} = V[bh, \text{Block}_N]$ \Comment{$[\text{Block}_N, D]$, stays in registers}
    \Statex \quad \textit{// Parent assignment (tiled over $M_0$ if codebook exceeds SRAM)}
    \State $\mathbf{c}_p = C[h,\; 0\!:\!M_0]$ \Comment{$[M_0, D]$, loaded into SRAM}
    \State $\mathbf{d} = \|\mathbf{k} - \mathbf{c}_p\|^2$ \Comment{$[\text{Block}_N, M_0]$ pairwise}
    \State $\mathbf{a} = \arg\min(\mathbf{d}, \text{axis}{=}1)$; \quad $\mathbf{d}_{\text{best}} = \min(\mathbf{d}, \text{axis}{=}1)$ \Comment{$[\text{Block}_N]$ each}
    \State $\bar{V}[bh, \mathbf{a}] \mathrel{+}= \mathbf{v}$; \quad $n[bh, \mathbf{a}] \mathrel{+}= 1$ \Comment{Atomic add}
    \Statex \quad \textit{// Child assignment --- children of parent $m$ at $C[h,\; M_0 + m\mathcal{C}\!:\!M_0 + (m{+}1)\mathcal{C}]$}
    \State $\mathbf{c}_0 = M_0 + \mathbf{a} \cdot \mathcal{C}$ \Comment{$[\text{Block}_N]$ first-child index per key}
    \State $\mathbf{d}_{\text{child}} = \text{full}(\text{Block}_N,\; \infty)$; \quad $\mathbf{c}^* = \mathbf{c}_0$
    \For{$c = 0, \ldots, \mathcal{C}-1$}
        \State $\mathbf{c}_{\text{code}} = C[h,\; \mathbf{c}_0 + c]$ \Comment{$[\text{Block}_N, D]$}
        \State $\mathbf{d}_c = \|\mathbf{k} - \mathbf{c}_{\text{code}}\|^2$ \Comment{$[\text{Block}_N]$}
        \State \textbf{where} $\mathbf{d}_c < \mathbf{d}_{\text{child}}$: \; $\mathbf{d}_{\text{child}} = \mathbf{d}_c$, \; $\mathbf{c}^* = \mathbf{c}_0 + c$
    \EndFor
    \State \textbf{where} $\mathbf{d}_{\text{child}} < \mathbf{d}_{\text{best}}$: \; $\bar{V}[bh, \mathbf{c}^*] \mathrel{+}= \mathbf{v}$; \quad $n[bh, \mathbf{c}^*] \mathrel{+}= 1$ \Comment{Masked atomic add}
\EndFor
\end{algorithmic}
\end{algorithm}
VQ-attention introduces several sequential operations --- quantization, value aggregation, and attention --- with intermediate results passing through global memory. While $\mathcal{O}(MN)$ complexity guarantees asymptotic efficiency, we improve wall-clock performance by fusing quantization and value aggregation into a single kernel that keeps each key in registers across both parent and child assignment. Each key's value is then scattered directly to its assigned codeword's accumulator via atomic operations at the tile level. The reduced memory traffic becomes increasingly important as $N$ and $M$ grow, and additionally allows VQ-attention to compete with Flash Attention already at moderate sequence lengths where the complexity gap alone is insufficient. While we present this for AVQ-attention, the fused VQ kernel is equally an improvement for standard VQ-attention. We provide pseudocode in \cref{alg:vqprecompute}.
\subsection{Computing (A)VQ-Attention with Flash Attention}
\paragraph{Tiled VQ-attention.}
The VQ-attention formulation maps directly to Flash Attention's tiling strategy. Continuing with the notation from \cref{sec:flashattention}, we replace $K_J$ with $C_J$, so that indices $J$ and $j$ now refer to codewords rather than keys. We write $\bar{V}_j = \sum_{k: a(K_k)=j} V_k$ for the aggregated values per codeword. The key difference from standard attention is that each codeword represents $n_j$ keys, so the denominator accumulates counts rather than row sums:

\begin{equation}
\label{eq:vqflashattention}
\begin{aligned}
A_{IJ} &= \exp(Q_I C_J^{\top}) \\
X_I(J) &= A_{IJ} \bar{V}_J \\
Z_I(J) &= A_{IJ} n_J \\
\bar{X}_I &= \sum_J X_I(J), \quad \bar{Z}_I = \sum_J Z_I(J) \\
Y_I &= \bar{X}_I \,/\, \bar{Z}_I
\end{aligned}
\end{equation}
where $n_j$ denotes the count of keys quantized to codeword $C_j$. The accumulators $\bar{X}_I$ and $\bar{Z}_I$ are built up incrementally across tiles using the online softmax trick for numerical stability, as in Flash Attention. We leave the subtraction of a stable maximum in the exponent implicit throughout the text; the pseudocode in \cref{alg:fusedattention} shows the full computation and \cref{app:effective_logits} discusses the choice of running maximum in the presence of codeword counts.
\paragraph{Codeword importance.}
\label{sec:importance}
We can use $A_{IJ}$ and the accumulated denominator $\bar{Z}_I$ to extract per-codeword importance scores. We define importance for each query tile $I$ as:
\begin{equation}
\label{eq:importance}
w_j(I) = \sum_{i \in I} \frac{A_{ij} \cdot n_j}{\bar{Z}_i}
\end{equation}
where $\bar{Z}_i$ is the $i$-th element of $\bar{Z}_I$. Importance is thus computed from the denominator that the online softmax accumulation already maintains, at minimal extra cost. When $M_0$ is small enough that all codewords fit in SRAM, $\bar{Z}_i$ is exact after a single pass; for larger $M_0$ requiring tiling over codewords, an approximate denominator can be used (see \cref{app:importance_tiling}).

\paragraph{Child spawning.}
For each query tile $I$, the top-$\mathcal{P}$ parents by importance are selected for refinement. For each selected parent, its $\mathcal{C}$ (contiguous) children are looked up in the codebook tree. Child aggregated values $\bar{V}_c$ and counts $n_c$ are immediately available, having been precomputed by the VQ kernel (\cref{alg:vqprecompute}). Together with the child codewords $C_{p,c}$ loaded from the codebook, this provides everything needed for child attention.

\paragraph{Child attention.}
\label{sec:child_attention}
When children are added, some keys shift from their originally assigned parent to a closer child codeword. The parent's aggregated values $\bar{V}_p$ and counts $n_p$ no longer fully represent these keys, so its contribution in the accumulators $\bar{X}_I$ and $\bar{Z}_I$ must be corrected. Crucially, we want to avoid recomputing the parent logits $S_{ip} := Q_i C_p^\top$ for two reasons: it would cost extra FLOPs, and it would require keeping parent codewords in SRAM while processing children. We structure the computation as a tiled Flash Attention pass where the first tile(s) consist of parents and subsequent tiles consist of children. To correct the parent contribution without revisiting it, we exploit the parent-child constraint (\cref{eq:parent_constraint}): since $C_p = \frac{1}{\mathcal{C}}\sum_c C_{p,c}$, we have
\begin{equation}
\label{eq:parent_recovery}
S_{ip} = Q_i C_p^\top = \frac{1}{\mathcal{C}}\sum_{c=1}^{\mathcal{C}} Q_i C_{p,c}^\top = \frac{1}{\mathcal{C}}\sum_{c=1}^{\mathcal{C}} S_{ic}
\end{equation}
so parent logits are recovered directly from the child logits already being computed. Since children are stored contiguously in the codebook, the sum in \cref{eq:parent_recovery} reduces over adjacent entries in a register tile and adds negligible cost. We then define the \emph{correcting attention}:
\begin{equation}
\label{eq:correcting_attention}
\Delta A_{ic} = \exp(S_{ic}) - \exp(S_{ip})
\end{equation}
Updating the accumulators with $\Delta A$ in place of $A$ implicitly removes the parent's contribution for keys that moved to children and replaces it with their respective child's attention weight (derivation in \cref{app:correcting_attention}). This requires only a single dot product per parent tile. Pseudocode for the full fused attention kernel is given in \cref{alg:fusedattention}.
\renewcommand{\algorithmicrequire}{\textbf{Input:}}
\renewcommand{\algorithmicensure}{\textbf{Output:}}
\begin{algorithm}[h!]
\caption{Flash AVQ-Attention kernel.}
\label{alg:fusedattention}
\begin{algorithmic}[1]
\Require $Q \in \mathbb{R}^{BH \times N \times D}$, \; $C \in \mathbb{R}^{H \times M_{\text{total}} \times D}$, \; $\bar{V} \in \mathbb{R}^{BH \times M_{\text{total}} \times D}$, \; $n \in \mathbb{R}^{BH \times M_{\text{total}}}$
\Ensure $Y \in \mathbb{R}^{BH \times N \times D}$
\For{each $(bh, \text{Block}_N)$ \textbf{in parallel}} \Comment{Separate GPU programs}
    \State $\mathbf{q} = Q[bh, \text{Block}_N]$ \Comment{$[\text{Block}_N, D]$, stays in registers}
    \State $\mathbf{c} = C[h,\; 0\!:\!M_0]$; \quad $\bar{\mathbf{v}} = \bar{V}[bh,\; 0\!:\!M_0]$; \quad $\mathbf{n} = n[bh,\; 0\!:\!M_0]$ \Comment{Into SRAM}
    \State $\mathbf{s} = \mathbf{q}\, \mathbf{c}^\top$ \Comment{$[\text{Block}_N, M_0]$ logits}
    \State $\mathbf{m} = \max_{j:\, \mathbf{n}_j > 0} \mathbf{s}_{:,j}$ \Comment{Stable max (empty codes excluded)}
    \State $\mathbf{A} = \exp(\mathbf{s} - \mathbf{m})$; \quad $\mathbf{A}_{:,j} = 0 \;\; \forall\, j\!: \mathbf{n}_j {=} 0$
    \State $\bar{\mathbf{x}} = \mathbf{A}\, \bar{\mathbf{v}}$; \quad $\bar{\mathbf{z}} = \textstyle\sum_j \mathbf{A}_{:,j} \cdot \mathbf{n}_j$
    \State $\mathbf{w}_j = \textstyle\sum_{i} \mathbf{A}_{ij}\, \mathbf{n}_j \,/\, \bar{\mathbf{z}}_i$ \Comment{$[M_0]$ importance (\cref{eq:importance})}
    \Statex \quad \textit{// Top-$\mathcal{P}$ selection and child refinement}
    \State $\mathcal{S} = \text{top-}\mathcal{P}(\mathbf{w})$ \Comment{Selected parent indices}
    \For{each selected parent $p \in \mathcal{S}$} \Comment{Optionally in $\lceil \mathcal{P}/\text{Block}_\mathcal{P} \rceil$ tiles}
        \State Load $\mathbf{c}_c,\, \bar{\mathbf{v}}_c,\, \mathbf{n}_c$ for children of $p$
        \State $\mathbf{s}_c = \mathbf{q}\, \mathbf{c}_c^\top$ \Comment{$[\text{Block}_N, \mathcal{C}]$ child logits}
        \Statex \quad\quad \textit{// Online softmax merge}
        \State $\mathbf{m}' = \max\!\big(\mathbf{m},\; \max_{c:\, \mathbf{n}_c > 0} \mathbf{s}_{:,c}\big)$
        \State $\bar{\mathbf{x}} = \bar{\mathbf{x}} \cdot \exp(\mathbf{m} - \mathbf{m}')$; \quad $\bar{\mathbf{z}} = \bar{\mathbf{z}} \cdot \exp(\mathbf{m} - \mathbf{m}')$; \quad $\mathbf{m} = \mathbf{m}'$
        \Statex \quad\quad \textit{// Parent correction via \cref{eq:parent_recovery}}
        \State $\mathbf{s}_p = \frac{1}{\mathcal{C}} \textstyle\sum_c \mathbf{s}_c$ \Comment{Recover parent logits from children}
        \State $\mathbf{A}_c = \exp(\mathbf{s}_c - \mathbf{m})$; \quad $\mathbf{A}_p = \exp(\mathbf{s}_p - \mathbf{m})$
        \State $\Delta\!\mathbf{A} = \mathbf{A}_c - \mathbf{A}_p$; \quad $\Delta\!\mathbf{A}_{:,c} = 0 \;\; \forall\, c\!: \mathbf{n}_c {=} 0$
        \State $\bar{\mathbf{x}} \mathrel{+}= \Delta\!\mathbf{A}\; \bar{\mathbf{v}}_c$; \quad $\bar{\mathbf{z}} \mathrel{+}= \textstyle\sum_c \Delta\!\mathbf{A}_{:,c} \cdot \mathbf{n}_c$
    \EndFor
    \State $Y[bh, \text{Block}_N] = \bar{\mathbf{x}} \,/\, \bar{\mathbf{z}}$
\EndFor
\end{algorithmic}
\end{algorithm}
\paragraph{Summary.}
The full AVQ-attention forward pass consists of two fused kernels (\cref{fig:pipeline}): VQ Precompute (\cref{alg:vqprecompute}) and Flash Attention (\cref{alg:fusedattention}). \Cref{tab:complexity} compares per-step costs with flat VQ-attention.
\begin{figure}[t]
\centering
\input{figures/fig_pipeline.tex}
\caption{AVQ-Attention inference pipeline. \textbf{Kernel~1} (VQ Precompute): keys and values are quantized against the parent codebook $C_p$, producing aggregated values $\bar{V}_p$ and counts $n_p$. Child quantization reuses parent assignments~\textcircled{\raisebox{-0.5pt}{\scriptsize 1}}, so each key is compared only against the $\mathcal{C}$ children of its assigned parent. \textbf{Kernel~2} (Flash Attention): $\text{Attn}_p$ computes attention over parent codewords and extracts importance. The online softmax accumulators and per-tile importance scores are carried forward~\textcircled{\raisebox{-0.5pt}{\scriptsize 2}} to $\text{Attn}_c$, which refines the top-$\mathcal{P}$ most important parents with child attention using correcting attention weights.}
\label{fig:pipeline}
\end{figure}
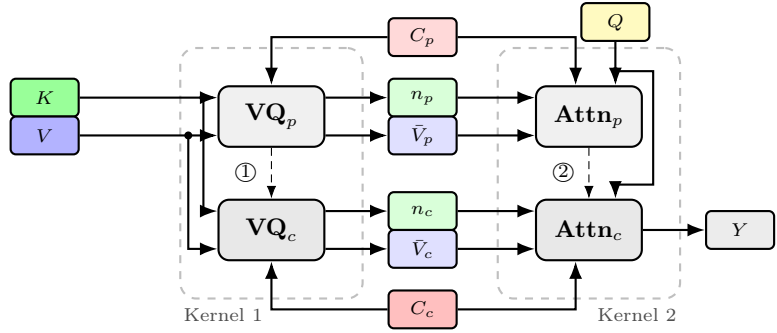
\begin{table}[h]
\centering
\caption{Complexity comparison: flat VQ-attention ($M$ codewords) vs.\ AVQ-attention ($M_0$ parents, $\mathcal{C}$ children per parent, $\mathcal{P}$ parents refined).}
\label{tab:complexity}
\begin{tabular}{lcc}
\toprule
\textbf{Step} & \textbf{VQ-Attention} & \textbf{AVQ-Attention} \\
\midrule
VQ assignment & $\mathcal{O}(NMD)$ & $\mathcal{O}(N(M_0 + \mathcal{C})D)$ \\
Value aggregation & $\mathcal{O}(ND)$ & $\mathcal{O}(ND)$ \\
(Parent) Attention & $\mathcal{O}(NMD)$ & $\mathcal{O}(NM_0 D)$ \\
Child attention & --- & $\mathcal{O}(N\mathcal{P}\mathcal{C}D)$ \\
\midrule
\textbf{Total FLOPs} & $\mathcal{O}(NMD)$ & $\mathcal{O}(N(M_0 + \mathcal{P}\mathcal{C})D)$ \\
\textbf{Codebook resolution} & $M$ & $M_0(1 + \mathcal{C})$ \\
\bottomrule
\end{tabular}
\end{table}

%% file: figures/fig_pipeline.tex
\begin{tikzpicture}[
    >=Latex,
    tensor/.style={draw, minimum width=0.9cm, minimum height=0.5cm,
                   font=\scriptsize, rounded corners=2pt, thick},
    block/.style={draw, minimum width=1.4cm, minimum height=0.8cm,
                  font=\small\bfseries, rounded corners=4pt, thick,
                  align=center},
    arr/.style={->, thick, >=Latex},
    darr/.style={->, >=Latex, densely dashed},
    dbox/.style={draw, gray!50, thick, rounded corners=6pt, dashed},
    dlbl/.style={font=\scriptsize, text=gray!60!black},
]

\node[tensor, fill=red!15] (Cp) at (5.0, 1.8) {$C_p$};

\node[tensor, fill=green!40] (K) at (0, 1.0) {$K$};
\node[tensor, fill=blue!30] (V) at (0, 0.5) {$V$};
\node[block, fill=gray!12] (VQp) at (3.0, 0.75) {$\text{VQ}_p$};
\node[tensor, fill=green!15] (np) at (5.0, 1.0) {$n_p$};
\node[tensor, fill=blue!12] (Vbarp) at (5.0, 0.5) {$\bar{V}_p$};
\node[block, fill=gray!12] (Attnp) at (7.2, 0.75) {$\text{Attn}_p$};

\node[block, fill=gray!18] (VQc) at (3.0, -0.75) {$\text{VQ}_c$};
\node[tensor, fill=green!15] (nc) at (5.0, -0.5) {$n_c$};
\node[tensor, fill=blue!12] (Vbarc) at (5.0, -1.0) {$\bar{V}_c$};
\node[block, fill=gray!18] (Attnc) at (7.2, -0.75) {$\text{Attn}_c$};

\node[tensor, fill=red!25] (Cc) at (5.0, -1.8) {$C_c$};

\node[tensor, fill=gray!15] (Y) at (9.2, -0.75) {$Y$};
\node[tensor, fill=yellow!30] (Q) at (7.55, 2.0) {$Q$};

\node[dbox, fit=(VQp)(VQc), inner sep=14pt] (k1box) {};
\node[dlbl, anchor=north east] at (k1box.south) {Kernel 1};
\node[dbox, fit=(Attnp)(Attnc), inner sep=14pt] (k2box) {};
\node[dlbl, anchor=north west] at (k2box.south) {Kernel 2};

\draw[arr] (K.east) -- (VQp.west |- K);
\draw[arr] (V.east) -- (VQp.west |- V);
\draw[arr] (VQp.east |- K) -- (np.west);
\draw[arr] (np.east) -- (Attnp.west |- np);
\draw[arr] (VQp.east |- V) -- (Vbarp.west);
\draw[arr] (Vbarp.east) -- (Attnp.west |- Vbarp);

\coordinate (Vsplit) at (1.9, 0.5);
\fill (Vsplit) circle (1.5pt);
\draw[arr] (Vsplit) |- (VQc.west |- Vbarc);
\coordinate (Ksplit) at (2.1, 1.0);
\fill (Ksplit) circle (1.5pt);
\draw[arr] (Ksplit) |- (VQc.west |- nc);

\draw[arr] (VQc.east |- nc) -- (nc.west);
\draw[arr] (nc.east) -- (Attnc.west |- nc);
\draw[arr] (VQc.east |- Vbarc) -- (Vbarc.west);
\draw[arr] (Vbarc.east) -- (Attnc.west |- Vbarc);
\draw[arr] (Attnc.east) -- (Y.west);

\draw[darr] (VQp.south) -- (VQc.north)
    node[midway, left=2pt, font=\scriptsize] {\textcircled{\raisebox{-0.5pt}{\scriptsize 1}}};
\draw[darr] (Attnp.south) -- (Attnc.north)
    node[midway, left=2pt, font=\scriptsize] {\textcircled{\raisebox{-0.5pt}{\scriptsize 2}}};

\draw[arr] (Cp.west) -| (VQp.north);
\draw[arr] (Cp.east) -| ([xshift=-5pt]Attnp.north);

\draw[arr] (Cc.west) -| (VQc.south);
\draw[arr] (Cc.east) -| ([xshift=-5pt]Attnc.south);

\draw[thick] (Q) -- (7.55, 1.35);
\fill (7.55, 1.35) circle (1.5pt);
\draw[arr] (7.55, 1.35) -- ([xshift=10pt]Attnp.north);
\draw[arr] (7.55, 1.35) -- (8.05, 1.35) -- (8.05, -0.15) -- (7.55, -0.15) -- ([xshift=10pt]Attnc.north);

\end{tikzpicture}

%% file: sections/experiments.tex
\section{Experiments and Results}
\label{sec:experiments}
\begin{figure}[h!]
\centering
\includegraphics[width=\linewidth]{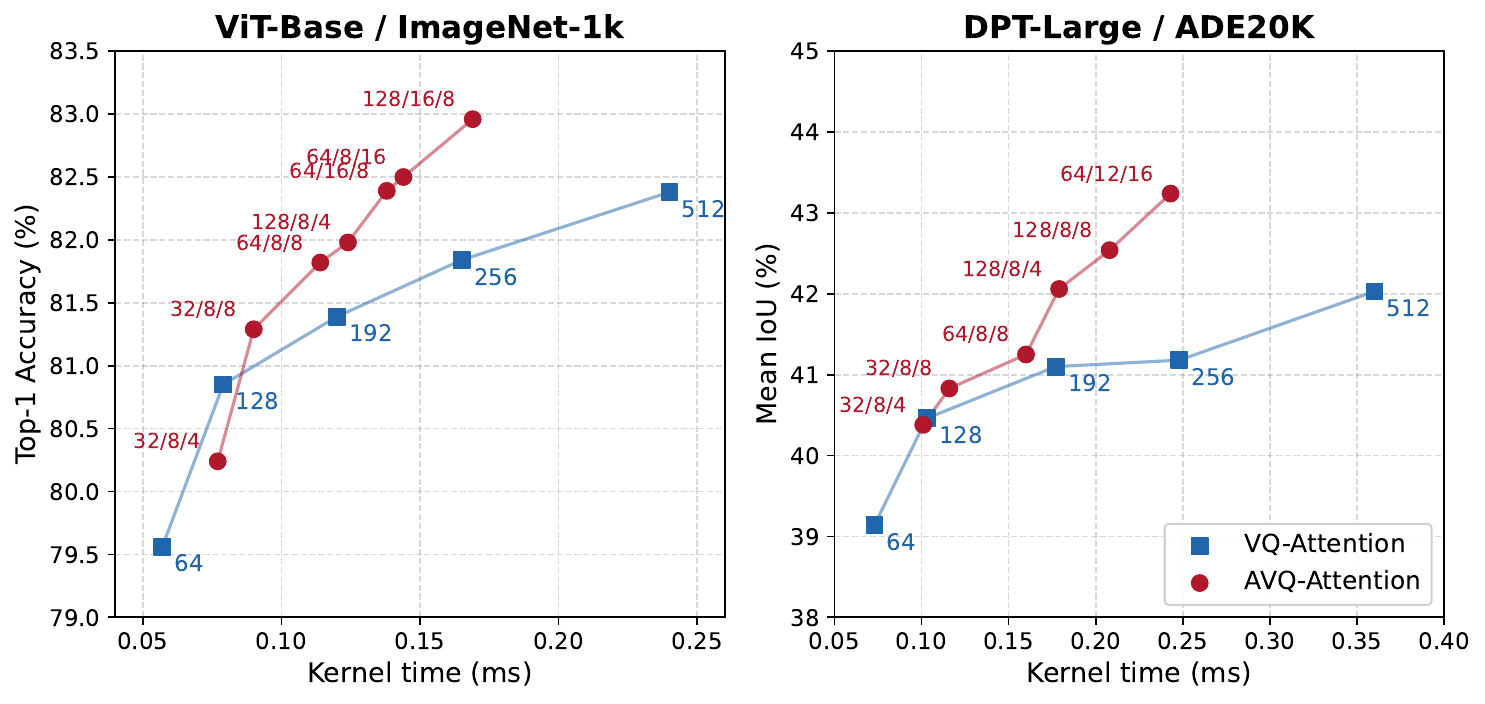}
\caption{Task performance vs.\ attention kernel time (ms) for VQ-attention (blue, labeled by $M$) and AVQ-attention (red, labeled by $M_0/\mathcal{P}/\mathcal{C}$).
\textbf{Left:} Top-1 accuracy on ImageNet-1k ($N{=}785$). \textbf{Right:} Mean IoU on ADE20K ($N{=}901$). AVQ achieves higher quality than VQ at comparable cost on both tasks.}
\label{fig:accuracy_cost}
\end{figure}

Starting from pretrained transformers, we replace attention layers with (A)VQ-attention and fine-tune for a small number of epochs. We evaluate on image classification (ImageNet-1k \cite{deng2009imagenet}) using a ViT-Base \cite{dosovitskiy2021vit} ($N{=}785$ tokens, 85.8\% top-1) and semantic segmentation (ADE20K \cite{zhou2017ade20k}) using DPT-Large \cite{ranftl2021dpt} ($N{=}901$ tokens, 49.0\% mIoU). Full training details are in \cref{app:training_details}. \Cref{fig:accuracy_cost} reports task performance vs.\ attention kernel time under identical training conditions. On both tasks, AVQ-attention consistently outperforms flat VQ-attention at comparable cost, confirming that adaptive codebook allocation improves the accuracy--efficiency trade-off. We further analyze codebook properties and attention-mass concentration in \cref{app:results_analysis}.
\subsection{Scaling Analysis}
\label{sec:exp_scaling}
The efficiency advantages of (A)VQ-attention grow with sequence length. To validate the complexity analysis from \cref{tab:complexity}, we benchmark wall-clock kernel time across sequence lengths; \cref{fig:speed_scaling} in the Appendix confirms the predicted linear scaling with both $N$ and the number of codewords (setup details in \cref{app:benchmark_details}). \Cref{tab:speed_scaling} reports absolute wall-clock times for the larger configurations most relevant to long-sequence deployment, including an unfused VQ baseline to isolate the speedup from fusing the VQ precompute step. At $N{=}65{,}536$, our fused AVQ-attention configurations are $81$--$127\times$ faster than Flash Attention. For flat VQ-attention, comparing against the unfused baseline shows that fusing the VQ precompute step alone provides a ${\sim}2\times$ speedup.
\begin{table}[h!]
\centering
\caption{Wall-clock kernel time (ms) vs.\ sequence length $N$ ($B{=}4,H{=}12$, $D{=}64$). For fair comparison, all methods use flash-attention-style kernels for the attention step; ``unfused VQ'' uses \texttt{torch.compile}'d PyTorch only for VQ precompute, so the speedup to our fused rows isolates the contribution of fusing the VQ precompute step. AVQ configs denoted $M_0/\mathcal{P}/\mathcal{C}$. Setup details in \cref{app:benchmark_details}.}
\label{tab:speed_scaling}
\begin{tabular}{lrrrr}
\toprule
\textbf{Configuration} & \multicolumn{1}{c}{$N{=}$1k} & \multicolumn{1}{c}{$N{=}$4k} & \multicolumn{1}{c}{$N{=}$16k} & \multicolumn{1}{c}{$N{=}$64k} \\
\midrule
\multicolumn{5}{l}{\textit{Baselines}} \\
Flash-Attn ($\mathcal{O}(N^2)$) & 0.21 & 3.11  & 49.6  & 847   \\
VQ-attn (unfused VQ) $M{=}256$ & 0.35 & 1.29  & 5.16  & 20.7  \\
VQ-attn (unfused VQ) $M{=}512$ & 0.61 & 2.31  & 9.28  & 37.4  \\
\midrule
\multicolumn{5}{l}{\textit{Ours (fused kernels)}} \\
VQ-attn $M{=}256$      & 0.18 & 0.66  & 2.62  & 10.6  \\
VQ-attn $M{=}512$      & 0.30 & 1.10  & 4.40  & 17.9  \\
AVQ-attn $64/8/8$      & 0.13 & 0.43  & 1.65  & 6.67  \\
AVQ-attn $64/16/8$     & 0.16 & 0.51  & 1.99  & 7.95  \\
AVQ-attn $128/8/8$     & 0.17 & 0.60  & 2.37  & 9.46  \\
AVQ-attn $128/16/8$    & 0.21 & 0.67  & 2.59  & 10.4  \\
\bottomrule
\end{tabular}
\end{table}

\subsection{Elastic Inference via Top-$\mathcal{P}$ Adjustment}
Since $\mathcal{P}$ can be changed at inference time without retraining, a single model can trade speed for accuracy. \Cref{tab:elastic_p} shows this for $M_0{=}64$, $\mathcal{C}{=}8$, trained with full spawning ($\mathcal{P}{=}M_0$). We also report capture: the fraction of true attention weight falling on keys assigned to the selected $\mathcal{P}$ parents. At $\mathcal{P}{=}16$, capture exceeds 82\% on both tasks and performance is within 0.3\% of the maximum, explaining the diminishing returns at higher $\mathcal{P}$.

\subsection{Comparison Across Efficient-Attention Methods}
\label{sec:exp_baselines}
We compare against several representative efficient-attention methods (\cref{tab:ade20k_baselines}). AVQ reaches the highest mIoU at comparable kernel cost, confirming it is competitive as a drop-in replacement for the attention layer. 

\begin{table}[h!]
\centering
\setlength{\tabcolsep}{6pt}
\caption{ADE20K (DPT-L); mIoU $\pm$ std from 3 seeds. All methods use an identical 30-epoch recipe to give each the chance to saturate (\cref{app:training_details}). $^\dagger$not FA-compatible.}
\label{tab:ade20k_baselines}
\begin{tabular}{lcc}
\toprule
\textbf{Method} & \textbf{mIoU (\%)} & \textbf{Kernel (ms)} \\
\midrule
Flash Attention-v2 (baseline)              & 49.0  & 0.313 \\
\midrule
AVQ ($M_0{=}32, \mathcal{P}{=}8, \mathcal{C}{=}8$) & \textbf{43.33} $\pm$ 0.12 & 0.116 \\
Flat VQ ($M{=}128$)             & 42.70 $\pm$ 0.08 & 0.103 \\
Flat VQ ($M{=}192$)             & 43.04 $\pm$ 0.05 & 0.164 \\
Swin (window 7)~\cite{liu2021swin}      & 42.90 $\pm$ 0.09 & 0.108 \\
NATTEN (window 5)~\cite{hassani2023nat} & 40.99 $\pm$ 0.21 & 0.126 \\
Linformer ($k{=}64$)~\cite{wang2020linformer} & 38.80 & 0.139 \\
Performer ($r{=}256$)~\cite{choromanski2021performer} & 25.29 & 0.54$^\dagger$ \\
\bottomrule
\end{tabular}
\end{table}

\subsection{High-Resolution Diffusion}
\label{sec:exp_diffusion}
We evaluate AVQ-attention in the Stable Diffusion~1.5 (SD1.5) UNet~\cite{rombach2022ldm} following the distillation setup of LinFusion~\cite{liu2024linfusion}. We adopt the same distillation objective, data, and hyperparameters, training for 50k steps (half of their 100k). As the UNet's inner-most self-attention blocks operate on only $N{\le}256$ tokens, exact attention is already trivially cheap and a marginal part of the cost, leaving little to gain from replacing it. We therefore apply AVQ only to the five outer (level-0) blocks ($N{=}4096$) and for one experiment also to the five level-1 blocks ($N{=}1024$).
Using ScaleCrafter~\cite{he2024scalecrafter} we further test the models on unseen $1024^2$ inference resolutions, making AVQ-attention handle token numbers up to $N{\approx}16$k. 
We report FID~\cite{heusel2017fid} and CLIP score~\cite{radford2021clip} on COCO~\cite{lin2014coco} under LinFusion's evaluation protocol in \Cref{tab:sd_baselines}.
\begin{table}[t!]
\centering
\footnotesize
\setlength{\tabcolsep}{4pt}
\caption{SD1.5 on COCO under the LinFusion protocol; FID, CLIP, and UNet forward time per denoising step. $^\dagger$additionally replaces the five level-1 ($N{=}1$k) blocks.}
\label{tab:sd_baselines}
\begin{tabular}{lcccccc}
\toprule
 & \multicolumn{3}{c}{\textbf{512$^2$} (train)} & \multicolumn{3}{c}{\textbf{1024$^2$} (w/ ScaleCrafter)} \\
\cmidrule(lr){2-4} \cmidrule(lr){5-7}
\textbf{Method} & FID$\downarrow$ & CLIP$\uparrow$ & ms & FID$\downarrow$ & CLIP$\uparrow$ & ms \\
\midrule
SD1.5 (FA-v2)             & 12.75 & 0.318 & 44.37 & 41.03 & 0.292 & 213.43 \\
LinFusion (Mamba)         & 12.52 & 0.318 & 43.70 & 36.37 & 0.295 & 141.57 \\
ToMe-SD ($r{=}0.5$)       & \textbf{12.40} & 0.317 & 41.99 & 43.00 & 0.290 & 163.30 \\
\midrule
\multicolumn{7}{l}{\textbf{AVQ (ours)}} \\
\quad M32P8C8             & 12.50 & 0.319 & \textbf{39.09} & \textbf{35.39} & \textbf{0.297} & 133.48 \\
\quad M64P16C8            & 12.51 & \textbf{0.320} & 39.50 & 35.65 & 0.296 & 135.06 \\
\quad M128P32C8           & 12.55 & 0.319 & 40.23 & 35.51 & \textbf{0.297} & 138.16 \\
\quad M64P16C8 ($+$L1)$^\dagger$ & 12.48 & \textbf{0.320} & 39.41 & 36.29 & \textbf{0.297} & \textbf{128.59} \\
\bottomrule
\end{tabular}
\end{table}
\vspace{-1.25em}

%% file: sections/conclusion.tex
\section{Discussion and Conclusion}
\label{sec:conclusion}
We have presented AVQ-attention, an adaptive extension of vector-quantized attention that dynamically allocates codebook capacity to regions of key space receiving high attention mass. By combining a hierarchical codebook with importance-driven refinement, AVQ-attention achieves better task performance than flat VQ-attention at comparable cost.
The experiments validate AVQ-attention as a general-purpose attention mechanism rather than targeting benchmark-specific performance: we take pretrained transformers, replace their attention layers, and fine-tune for only a few epochs. Our focus was on the controlled comparison between AVQ, VQ and other efficient attention types, and given the promise of (A)VQ-attention as a competitive layer type, we leave the design of efficient end-to-end transformer architectures around it as future research.
Some training choices remain lightly explored.  For instance, the optimal $M_0$, $\mathcal{P}$, and $\mathcal{C}$ likely vary across layers --- some layers may need far fewer codewords (\cref{app:perlayer_analysis}) --- making per-layer configuration an attractive direction. AVQ-attention facilitates this by exposing informative measures such as captured attention mass and quantization error. The latter could additionally serve as a refinement criterion alongside importance, prioritizing parents whose children differ most from the parent representation. Together, these signals could enable efficient architecture search without full retraining.
Beyond per-layer tuning, deeper hierarchies are a natural next step: each additional depth adds only $\mathcal{P}\mathcal{C}$ codes to the attention cost while multiplying codebook resolution by $\mathcal{C}$, potentially yielding exponential resolution growth for linear cost.

%% file: sections/appendix.tex
\section{Codeword Clustering During Training}
\label{app:codewordtraining}
During training, we maintain codebook positions through exponential moving averages (EMA) while enforcing the parent-child constraint $C_p = \frac{1}{\mathcal{C}}\sum_{c=1}^{\mathcal{C}} C_{p,c}$. This section describes how we update codewords when new data arrives at time $t+1$.

\paragraph{Dual representation for codewords.} Let $p$ be a parent with $\mathcal{C}$ children. For each child $c$, we maintain two representations.
\begin{itemize}
    \item \textbf{Unconstrained EMA statistics}: $(S_c, N_c)$ accumulate keys assigned to child $c$ via standard EMA, tracking where it would naturally cluster without constraints. These define unconstrained means $M_c = S_c / N_c$.
    \item \textbf{Constrained positions}: $C_{p,c}$ are the actual codeword positions used for quantization and attention, derived from the unconstrained statistics while satisfying the constraint $C_p = \frac{1}{\mathcal{C}}\sum_c C_{p,c}$.
\end{itemize}
The unconstrained statistics $(S_c, N_c)$ preserve the full EMA history of where data naturally lies, while the constrained positions enforce the geometric relationship with the parent.

\paragraph{Parent update and child adjustment.} When new keys arrive at time $t+1$, we update each code's EMA statistics:
\begin{align}
S_x^{(t+1)} &= \lambda S_x^{(t)} + (1-\lambda) \sum_{k: \hat{k}=C^{(t)}_x} k \\
N_x^{(t+1)} &= \lambda N_x^{(t)} + (1-\lambda) |\{k :\hat{k}=C^{(t)}_x\}|
\end{align}
where $\lambda$ is the EMA decay rate. For parents we use no constraints and set $C_p^{(t+1)} = M_p^{(t+1)}= S_p^{(t+1)}/N_p^{(t+1)}$ directly. However, updating children using the same direct $C_c^{(t+1)}=M_c^{(t+1)}$ would violate the parent-child constraint.
To restore the constraint with minimal disruption, we solve the mass-weighted least-squares problem:
\begin{align*}
\min_{C_{p,1}, \ldots, C_{p,\mathcal{C}}} &\sum_{c=1}^{\mathcal{C}} N_c \|M_c - C_{p,c}\|^2 \\
\text{subject to:} \quad & \frac{1}{\mathcal{C}}\sum_{c=1}^{\mathcal{C}} C_{p,c} = C_p
\end{align*}
Via Lagrange multipliers, this yields a closed-form projection. Defining the constraint residual $\delta^{(t+1)} = \sum_c M_c^{(t+1)} - \mathcal{C}\, C_p^{(t+1)}$ and the harmonic weight $\sigma^{(t+1)} = \sum_c 1/N_c^{(t+1)}$:
\begin{equation}
\label{eq:child_constraint_solve}
C_{p,c}^{(t+1)} = M_c^{(t+1)} - \frac{\delta^{(t+1)}}{N_c^{(t+1)} \cdot \sigma^{(t+1)}}
\end{equation}
Each child is pulled toward the constraint surface by an amount inversely proportional to its mass---heavier children (larger $N_c$) resist displacement more.
These adjusted positions are used when children receive new data in the next step, while maintaining the constraint with the updated parent.

\section{Computing Importance with Tiled Codewords}
\label{app:importance_tiling}

When $M_0$ is too large for all codewords to fit in SRAM, the attention pass must tile over codewords. Computing importance $w_j$ (\cref{eq:importance}) in this setting presents a challenge: the denominator $\bar{Z}_i = \sum_{j'} A_{ij'} n_{j'}$ is only available after processing all codeword tiles, but we want to extract importance within each tile to avoid recomputing logits, and reduce it across queries within each tile to save memory.

We address this by using an \emph{approximate} denominator based on count-based extrapolation. Let $t$ denote a tile index set for codewords, with $t \leq J$ meaning tile $t$ is processed before or at tile $J$. After processing tiles up to $J$, the partial denominator $\sum_{t \leq J} Z_I(t)$ accounts for only those keys whose parent codewords have been visited. Let $N_{\text{seen}}(J) = \sum_{t \leq J} \sum_{j \in t} n_j$ be the total number of keys whose codewords have been processed so far. We extrapolate to the full key count:
\begin{equation}
\label{eq:aproxdenom}
\tilde{Z}_I(J) := \frac{N}{N_{\text{seen}}(J)} \sum_{t \leq J} Z_I(t)
\end{equation}
where $N$ is the total number of keys. Codeword importance is then computed as in \cref{eq:importance}, but with the approximate denominator:
\begin{equation}
\label{eq:codewordimportance}
w_j(I) = \sum_{i \in I} \frac{A_{ij} \cdot n_j}{\tilde{Z}_i(J)}, \quad j \in J,
\end{equation}
where $J$ is the codeword tile containing $j$. After all codeword tiles are processed for a given query tile $I$, it has importance scores for all $M_0$ codewords and independently selects its top-$\mathcal{P}$ parents for refinement. In our experimental settings, all relevant codebook sizes already fit within a single SRAM tile. To evaluate the approximation nonetheless, we train AVQ models with $M_0{=}256$, $\mathcal{C}{=}8$ and force tiling into two tiles of $128$ codewords each at evaluation time. \Cref{tab:tiled_capture} reports the fraction of true attention mass captured by the top-$\mathcal{P}$ parents selected via three methods: tiled importance (using the approximate denominator \cref{eq:aproxdenom}), non-tiled importance (exact denominator from a single pass over all codewords), and exact selection (computing the full $N{\times}N$ attention matrix and selecting the $\mathcal{P}$ parents whose assigned keys receive the most true attention mass).

\begin{table}[h]
\centering
\caption{Top-$\mathcal{P}$ capture (\% of true attention mass) for $M_0{=}256$, $\mathcal{C}{=}8$ with tile size 128. Tiled importance uses the count-based extrapolated denominator; non-tiled uses the exact denominator; exact selects the $\mathcal{P}$ parents with highest true attention mass, computed from the full attention matrix.}
\label{tab:tiled_capture}
\begin{tabular}{llccc}
\toprule
\textbf{Dataset} & $\mathcal{P}$ & \textbf{Tiled} & \textbf{Non-tiled} & \textbf{Exact} \\
\midrule
ImageNet (ViT-Base)   & 24 & 71.6\% & 75.9\% & 80.1\% \\
ImageNet (ViT-Base)   & 32 & 77.7\% & 82.0\% & 85.5\% \\
ADE20K (DPT-Large)    & 24 & 80.0\% & 83.2\% & 84.3\% \\
ADE20K (DPT-Large)    & 32 & 85.9\% & 88.6\% & 89.4\% \\
\bottomrule
\end{tabular}
\end{table}

\section{Correcting Attention Derivation}
\label{app:correcting_attention}

We show that updating the online softmax accumulators with the correcting attention $\Delta A_{ic} = A_{ic} - A_{ip}$ correctly replaces the parent's contribution with finer-grained child contributions.

Let parent $p$ have $n_p$ assigned keys with aggregated values $\bar{V}_p$. After refinement, each child $c$ receives $n_c$ keys with aggregated values $\bar{V}_c$; the remaining keys stay with the parent:
\begin{equation}
\bar{V}_{\text{stay}} = \bar{V}_p - \sum_{c=1}^{\mathcal{C}} \bar{V}_c, \qquad n_{\text{stay}} = n_p - \sum_{c=1}^{\mathcal{C}} n_c.
\end{equation}

The initial attention pass included the parent's contribution $A_{ip}\, \bar{V}_p$ in the numerator accumulator. The correct post-refinement contribution is $A_{ip}\, \bar{V}_{\text{stay}} + \sum_c A_{ic}\, \bar{V}_c$, so the required correction is:
\begin{equation}
\sum_c A_{ic}\, \bar{V}_c + A_{ip}\, \bar{V}_{\text{stay}} - A_{ip}\, \bar{V}_p = \sum_c A_{ic}\, \bar{V}_c - A_{ip} \sum_c \bar{V}_c.
\end{equation}
Since $A_{ip}$ is constant across children, this factorizes into a single dot product:
\begin{equation}
\label{eq:correction_factorization}
\sum_c \big(A_{ic} - A_{ip}\big)\, \bar{V}_c = \sum_c \Delta A_{ic}\, \bar{V}_c
\end{equation}
The denominator correction $\sum_c \Delta A_{ic}\, n_c$ follows identically. Thus the full correction for both accumulators is computed with one dot product of $\Delta A$ against the child aggregates, without needing to load the parent codeword $C_p$, its aggregated values $\bar{V}_p$, or counts $n_p$, and without computing or storing the parent logits $Q_i C_p^\top$ separately.

\section{Numerical Stability of Online Softmax in VQ-Attention}
\label{app:effective_logits}

The online softmax maintains a running maximum $m_i$ and computes all attention weights as $\exp(S_{ij} - m_i)$. The choice of $m_i$ does not affect the final result (it cancels between numerator and denominator), but it controls numerical precision. In AVQ-attention, the refinement round introduces child codewords that may have zero assigned keys ($n_c = 0$). Since empty codewords are not constrained by any assigned keys, their logits $S_{ic} = Q_i C_c^\top$ can lie far from the populated region of key space. If such a codeword sets the running maximum to a value much larger than the previous maximum, the rescaling factor $\exp(m_{\text{old}} - m_{\text{new}})$ could underflow to zero, destroying all previously accumulated attention information --- even though the empty codeword contributes nothing to the output ($\bar{V}_c = 0$, $n_c = 0$).

A natural idea is to redefine the logits as $S_{ij}' = Q_i C_j^\top + \ln n_j$, so that empty codewords ($n_j = 0$, $\ln 0 = -\infty$) are automatically excluded from the maximum. However, this complicates the parent logit recovery in \cref{eq:parent_recovery}: without folded counts, parent logits are recovered as a simple average of child logits $S_{ip} = \frac{1}{\mathcal{C}}\sum_c S_{ic}$, which reduces over adjacent register entries at negligible cost. With modified logits, the counts must first be subtracted out: $S_{ip} = \frac{1}{\mathcal{C}}\sum_c (S'_{ic} - \ln n_c) + \ln n_p$, requiring per-child counts alongside the logits.

A simpler solution is to restrict the running maximum to non-empty codewords:
\begin{equation}
m_i = \max_{j:\, n_j > 0} S_{ij}
\end{equation}
This requires only a masked comparison (one bit per codeword) rather than an extra floating-point value per codeword. Empty codewords contribute nothing (both $\bar{V}_j = 0$ and $n_j = 0$), but their attention weights $\exp(S_{ij} - m_i)$ may overflow in reduced precision since $S_{ij}$ can exceed $m_i$. We therefore zero out $A_{ij}$ for empty codewords in the parent pass and $\Delta A_{ic}$ for empty children in the refinement pass, preventing floating-point artifacts ($\infty \times 0 = \text{NaN}$).

\section{Benchmark Details}
\label{app:benchmark_details}

All wall-clock kernel times reported in the paper (\cref{fig:accuracy_cost}, \cref{tab:speed_scaling}, \cref{tab:elastic_p}, and \cref{tab:ade20k_baselines}) are measured on a single NVIDIA RTX~3090 (Ampere, SM~8.6, 24\,GB) at batch size $B{=}4$ and head dimension $D{=}64$, with $H{=}12$ heads for ViT-Base/ImageNet and $H{=}16$ for DPT-Large/ADE20K. The synthetic-length sweep in \cref{tab:speed_scaling} uses $N \in \{1024, 4096, 16384, 65536\}$ ($H{=}12$). VQ and AVQ timings include both the VQ precompute kernel and the attention kernel, launched via our fused two-kernel pipeline. Flash Attention uses PyTorch's \texttt{scaled\_dot\_product\_attention} with the Flash Attention 2 backend. These kernel timings use Triton's \texttt{do\_bench} utility (100\,ms warmup, 200\,ms measurement window, median reported), averaged over 3 independent runs. The per-denoising-step UNet times in \cref{tab:sd_baselines} instead time the full UNet forward with CUDA graphs, on the same RTX~3090 at batch size $B{=}2$.

\begin{figure}[ht!]
\centering
\includegraphics[width=\linewidth]{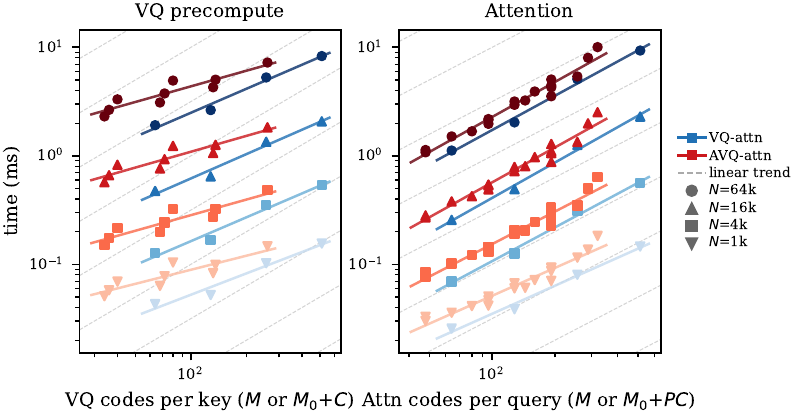}
\caption{Empirical verification of the complexity analysis in \cref{tab:complexity}. Kernel time vs.\ effective codebook size on log-log axes for VQ-attention (blue) and AVQ-attention (red), at four sequence lengths $N$ (light to dark). Gray dashed lines indicate slope~1 (linear scaling). All kernels scale linearly with $N$ (equidistant lines on the log scale for equal $4\times$ increases in $N$) and linearly with the number of codes, confirming $\mathcal{O}((M_0{+}\mathcal{C})ND)$ for VQ precompute and $\mathcal{O}((M_0{+}\mathcal{P}\mathcal{C})ND)$ for attention. The precompute kernels appear slightly sub-linear in the number of codes; this is a fixed per-kernel overhead (launch cost, memory setup) that is amortized as the codebook grows.}
\label{fig:speed_scaling}
\end{figure}

\Cref{fig:speed_scaling} plots our VQ and AVQ kernel times against effective codebook size at each $N$, empirically confirming the predicted linear scaling with both $N$ and codebook size.

\section{Training Details}
\label{app:training_details}

\paragraph{ImageNet-1k.} We use a ViT-Base backbone (patch size 8, $224{\times}224$ input, $N{=}785$ tokens including class token) pretrained on ImageNet-21k \cite{dosovitskiy2021vit}. All 12 attention layers are replaced with (A)VQ-attention. We fine-tune for 4 epochs using AdamW with learning rate $10^{-5}$, constant schedule with 1 epoch linear warmup, per-device batch size 64, and FP16 mixed precision. Data augmentation follows standard practice: RandomResizedCrop and RandomHorizontalFlip for training, Resize and CenterCrop for evaluation. Learnable key normalization (LayerNorm) stabilizes the key space and improves codebook training. The EMA decay is $\lambda = 0.99$ and the commitment loss weight is $\beta = 0.25$. For flat VQ-attention, we sweep codebook sizes $M \in \{64, 128, 256, 512\}$. For AVQ-attention, children are initialized at scale $0.1$ around their parent (i.e., $C_{p,c} = C_p + 0.1 \cdot \epsilon$, $\epsilon \sim \mathcal{N}(0, I)$). Queries are reordered along a Gilbert space-filling curve for spatial tile locality.

\paragraph{ADE20K.} We use DPT-Large \cite{ranftl2021dpt} (pretrained on ADE20K, $480{\times}480$ input, $N{=}901$ tokens) and replace all attention layers. We fine-tune for 10 epochs using AdamW with learning rate $10^{-5}$, constant schedule with 10\% linear warmup, per-device batch size 8, and FP16 mixed precision. The EMA decay warms up from $0.9$ to $0.99$ over the first 2 epochs (cosine schedule). All other VQ hyperparameters match the ImageNet setting. We select the best checkpoint by validation mIoU.

\paragraph{Efficient-attention comparison.} For \cref{tab:ade20k_baselines}, all methods---including (A)VQ---use a common 30-epoch ADE20K recipe, otherwise identical to the above. The longer schedule, together with dead-code handling for VQ (\cref{app:alt_training}), lifts the (A)VQ numbers relative to the 10-epoch main experiments, so the values in \cref{tab:ade20k_baselines} form a separate controlled comparison and are not directly comparable to \cref{fig:accuracy_cost}.

\section{Alternative VQ Training Methods}
\label{app:alt_training}
We learn the codebook with EMA-based online $k$-means (\cref{sec:method}), a simple choice that we do not claim to be optimal. Codebook learning is largely orthogonal to the attention mechanism, so improvements developed for vector quantization in other settings transfer directly to (A)VQ-attention. We briefly explored replacing the straight-through estimator with a differentiable quantizer (DiVeQ~\cite{vali2026diveq}) and per-batch dead-code handling that reassigns nearest keys to underused codewords. Both improved results in our setting. We nonetheless report the simpler EMA recipe throughout for consistency; we expect the accuracy--efficiency trade-offs we present to be conservative, with room for orthogonal advances in vector quantization to improve them further. 

\section{Results Analysis}
\label{app:results_analysis}

Below we report several supporting analyses of the trained AVQ models.

\paragraph{Elastic inference.} \Cref{tab:elastic_p} reports task performance, top-$\mathcal{P}$ capture, and kernel time for a single AVQ model ($M_0{=}64$, $\mathcal{C}{=}8$) trained at $\mathcal{P}{=}M_0$ and evaluated across a range of $\mathcal{P}$.

\begin{table}[t]
\centering
\caption{Elastic inference ($M_0{=}64$, $\mathcal{C}{=}8$): a single model trained with $\mathcal{P}{=}M_0$ evaluated at varying $\mathcal{P}$. Capture denotes the fraction of true attention weight on keys assigned to the selected $\mathcal{P}$ parents.}
\label{tab:elastic_p}
\begin{tabular}{lcccccc}
\toprule
Eval $\mathcal{P}$ & 4 & 8 & 12 & 16 & 24 & 32 \\
\midrule
\multicolumn{7}{l}{\textit{ADE20K ($N{=}901$, $H{=}16$)}} \\
mIoU (\%)    & 41.44 & 42.24 & 42.33 & 42.42 & 42.49 & 42.48 \\
Capture (\%) & 49.7  & 68.0  & 78.5  & 85.3  & 93.1  & 97.0  \\
Kernel ms    & 0.14  & 0.16  & 0.18  & 0.20  & 0.24  & 0.28  \\
\midrule
\multicolumn{7}{l}{\textit{ImageNet ($N{=}785$, $H{=}12$)}} \\
Acc (\%)     & 79.23 & 81.16 & 81.69 & 82.04 & 82.22 & 82.29 \\
Capture (\%) & 47.5  & 65.4  & 75.8  & 82.7  & 91.0  & 95.6  \\
Kernel ms    & 0.10  & 0.11  & 0.13  & 0.14  & 0.17  & 0.20  \\
\bottomrule
\end{tabular}
\end{table}

\begin{figure}[t]
\centering
\includegraphics[width=\linewidth]{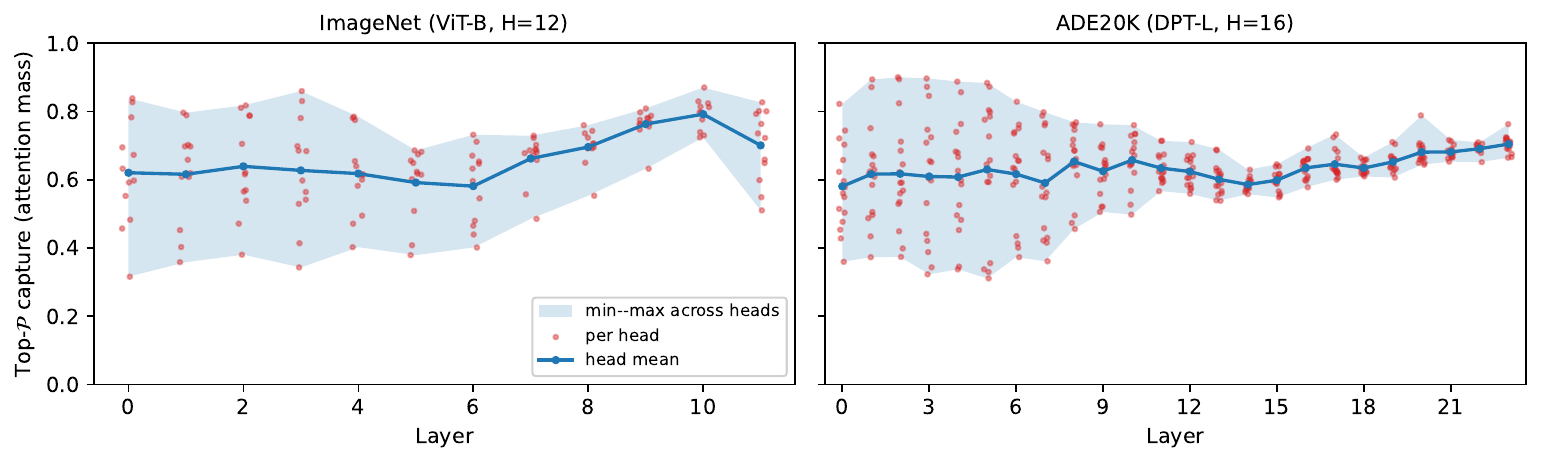}
\caption{Per-head top-$\mathcal{P}$ capture (fraction of true attention mass on
the $\mathcal{P}{=}8$ selected parents) for AVQ-attention ($M_0{=}64$,
$\mathcal{C}{=}8$), by layer: min--max across heads (shaded), individual heads
(points), and the head mean (line). \textbf{Left:} ImageNet (ViT-Base, $H{=}12$;
the final layer attends only from the CLS query). \textbf{Right:} ADE20K
(DPT-Large, $H{=}16$).}
\label{fig:perhead_capture}
\end{figure}

\paragraph{Per-head attention-mass concentration.}
\label{app:perhead_capture}
\Cref{fig:perhead_capture} reports top-$\mathcal{P}$ capture---the fraction of
true attention mass on the keys assigned to a head's $\mathcal{P}$ selected
parents---for every attention head of the trained AVQ model ($M_0{=}64$,
$\mathcal{C}{=}8$, evaluated at $\mathcal{P}{=}8$). Capture varies considerably
across heads within a layer on both backbones, and the head-mean is higher in
the later layers.

\paragraph{Per-layer codebook analysis.}
\label{app:perlayer_analysis}
All experiments in this paper apply the same codebook configuration ($M_0$, $\mathcal{C}$) uniformly across all attention layers. To investigate whether this is efficient, we run inference on our trained AVQ models and measure the per-layer commitment loss $\mathbb{E}\bigl[\lVert k - \hat{k} \rVert^2\bigr]$, the mean squared distance between each key and its assigned codeword. \Cref{tab:perlayer_imagenet} reports this for two codebook sizes on each dataset.

\begin{table}[h]
\centering
\caption{Per-layer commitment loss $\mathbb{E}[\lVert k - \hat{k}\rVert^2]$ for AVQ-attention ($\mathcal{C}{=}8$). Higher values may indicate that the codebook provides insufficient resolution for that layer's key distribution. This metric depends only on the codebook geometry and is independent of $\mathcal{P}$.}
\label{tab:perlayer_imagenet}
\begin{minipage}[t]{0.30\linewidth}
\centering
\subcaption*{\textit{ImageNet (ViT-Base)}}
\begin{tabular}{r cc}
\toprule
Layer & $M_0{=}64$ & $M_0{=}128$ \\
\midrule
0  &  8.0 &  7.2 \\
1  & 13.5 & 11.6 \\
2  & 23.1 & 23.0 \\
3  & 21.9 & 22.9 \\
4  & 23.8 & 25.0 \\
5  & 27.2 & 30.1 \\
6  & 24.4 & 27.8 \\
7  & 22.4 & 26.6 \\
8  & 19.8 & 24.1 \\
9  & 16.9 & 22.5 \\
10 & 14.6 & 21.2 \\
11 &  9.0 & 14.7 \\
\bottomrule
\end{tabular}
\end{minipage}%
\hfill
\begin{minipage}[t]{0.66\linewidth}
\centering
\subcaption*{\textit{ADE20K (DPT-Large)}}
\begin{tabular}{r cc @{\hskip 12pt} r cc}
\toprule
Layer & $M_0{=}64$ & $M_0{=}128$ & Layer & $M_0{=}64$ & $M_0{=}128$ \\
\midrule
0  &  6.9 &  4.7 & 12 & 11.5 &  4.7 \\
1  & 10.5 &  5.1 & 13 &  8.8 &  3.3 \\
2  & 15.0 &  7.3 & 14 &  6.5 &  2.3 \\
3  & 20.0 & 10.8 & 15 &  6.3 &  2.2 \\
4  & 24.3 & 14.6 & 16 &  6.7 &  2.4 \\
5  & 21.1 & 11.9 & 17 &  7.3 &  2.6 \\
6  & 22.0 & 12.4 & 18 &  6.0 &  2.0 \\
7  & 22.8 & 12.3 & 19 &  4.3 &  1.3 \\
8  & 21.8 & 12.5 & 20 &  4.6 &  1.3 \\
9  & 17.3 &  8.3 & 21 &  5.2 &  1.4 \\
10 & 18.1 &  9.4 & 22 &  6.2 &  1.7 \\
11 & 14.1 &  6.1 & 23 &  5.9 &  1.6 \\
\bottomrule
\end{tabular}
\end{minipage}
\end{table}

Commitment loss varies 3--4$\times$ across layers, with a consistent pattern across both datasets and codebook sizes: low in early layers, peaking in the middle layers (4--6), and decreasing again toward the end. In DPT-Large, the second half of the network (layers 12--23) has markedly lower commitment loss than the first half---below 5 throughout at $M_0{=}128$. Layer~0 combines low commitment loss with low codebook utilization (59--69\% active ratio at $M_0{=}64$).

\paragraph{Spatial reordering ablation.}
\label{app:gilbert_ablation}
As described in \cref{sec:method}, we reorder queries along a Gilbert space-filling curve so that contiguous tiles form spatially compact 2D regions. Since each tile independently selects which $\mathcal{P}$ parents to refine, spatial coherence ensures that queries sharing a refinement decision attend to similar parts of the image. We ablate this by evaluating a trained model ($M_0{=}64$, $\mathcal{P}{=}16$, $\mathcal{C}{=}8$) under three query orderings---\emph{Gilbert} (training configuration), \emph{raster} (no reordering, tiles are thin horizontal strips), and \emph{random} (destroys all locality)---and three tile sizes (\cref{tab:gilbert_ablation}).

On ADE20K, Gilbert reordering improves mIoU by ${\sim}0.2$ percentage points over raster order across all tile sizes; on ImageNet, Gilbert and raster perform identically (${\sim}82.0\%$). In the ImageNet runs the final layer computes attention only for the CLS query, so patch-token ordering does not affect its parent selection. Random permutation is clearly harmful in both settings ($-1.8\%$ accuracy on ImageNet, $-0.8\%$ mIoU on ADE20K). We observe no meaningful difference across tile sizes at the current sequence lengths.

\begin{table}[h]
\centering
\caption{Effect of query reordering on AVQ-attention ($M_0{=}64$, $\mathcal{P}{=}16$, $\mathcal{C}{=}8$) for ImageNet classification (accuracy~\%) and ADE20K segmentation (mIoU~\%). Random results averaged over 2 seeds.}
\label{tab:gilbert_ablation}
\begin{tabular}{lcccccc}
\toprule
 & \multicolumn{3}{c}{ImageNet (acc.~\%)} & \multicolumn{3}{c}{ADE20K (mIoU~\%)} \\
\cmidrule(lr){2-4}\cmidrule(lr){5-7}
\textbf{Ordering} & 32 & 64 & 128 & 32 & 64 & 128 \\
\midrule
Gilbert & 82.04 & 82.03 & 82.08 & 42.40 & 42.35 & 42.37 \\
Raster  & 82.07 & 82.00 & 82.02 & 42.16 & 42.20 & 42.15 \\
Random  & 80.18 & 80.20 & 80.22 & 41.37 & 41.39 & 41.40 \\
\bottomrule
\end{tabular}
\end{table}

\paragraph{Codebook utilization.}
\label{app:codebook_utilization}
A practical concern with VQ-attention is codebook utilization: as $M$ grows, some codewords attract few or no keys. We measure this via the \emph{active ratio}: the fraction of codewords whose assignment count over the full validation set exceeds 1\% of the fair-share count $N_{\text{val}}/M_{\text{total}}$ (e.g., $0.01 \times 50{,}000 \times 785 / 128 \approx 3{,}000$ for ImageNet at $M{=}128$), averaged across heads and layers.

\Cref{tab:active_ratio} reports active ratios for flat VQ-attention at varying $M$ and for AVQ-attention across several configurations, under identical training settings.

\begin{table}[h]
\centering
\caption{Codebook active ratio for flat VQ and AVQ under identical training settings. Flat VQ utilization degrades as $M$ grows, while AVQ maintains high utilization even at much larger total codebook sizes. The active ratio measures what fraction of codewords receive key assignments, forming the pool of codewords available for attention. In AVQ-attention, each query attends to only a subset of this pool, with different queries selecting different subsets.}
\label{tab:active_ratio}
\begin{tabular}{lcc}
\toprule
\textbf{Method} & $M_{\text{total}}$ & \textbf{Active ratio} \\
\midrule
Flat VQ, $M{=}64$   & 64   & 97.6\% \\
Flat VQ, $M{=}128$  & 128  & 95.7\% \\
Flat VQ, $M{=}256$  & 256  & 93.4\% \\
Flat VQ, $M{=}512$  & 512  & 90.4\% \\
\midrule
AVQ, $64/8/8$    & 576  & 97.0\% \\
AVQ, $64/16/8$   & 576  & 97.0\% \\
AVQ, $128/8/8$   & 1152 & 95.4\% \\
AVQ, $128/16/8$  & 1152 & 95.5\% \\
\bottomrule
\end{tabular}
\end{table}

Flat VQ exhibits a clear downward trend in utilization as $M$ grows: at $M{=}512$, nearly 10\% of codewords fall below the active threshold. AVQ maintains substantially higher utilization at much larger total codebook sizes: at $M_{\text{total}}{=}576$, AVQ achieves 97\% utilization, and even at $M_{\text{total}}{=}1152$ it retains over 95\%---better than flat VQ at $M{=}256$. The hierarchical structure naturally encourages full codebook usage: children are initialized near their parent, placing them in regions of key space where keys already concentrate.